\documentclass[pmlr]{jmlr}% new name PMLR (Proceedings of Machine Learning)

\usepackage[T1]{fontenc}

\usepackage{tgtermes}
\usepackage{amsmath}
\usepackage{scalefnt,letltxmacro}
\LetLtxMacro{\oldtextsc}{\textsc}
\renewcommand{\textsc}[1]{\oldtextsc{\scalefont{1.10}#1}}
\usepackage[scaled=0.92]{PTSans}
\usepackage{inconsolata}
\usepackage{mathbbol}
\usepackage{amssymb}

\usepackage{setspace}

\usepackage[final, expansion=alltext]{microtype}
\usepackage[english]{babel}
\usepackage{afterpage}
\usepackage{framed}
\usepackage{nicefrac}

{\endMakeFramed}

\renewcommand{\labelenumi}{\color{black!67}{\arabic{enumi}.}}

\usepackage{graphicx}
\usepackage[labelfont=bf]{caption}
\usepackage[format=hang]{subcaption}

\usepackage{booktabs}
\usepackage{multirow}
\usepackage{longtable}
\usepackage{etoolbox,siunitx}
\robustify\bfseries
\usepackage{algorithm2e}
\usepackage{listings}
\usepackage{fancyvrb}
\fvset{fontsize=\normalsize}
\usepackage{algorithm}
\usepackage{algorithmic}

\usepackage{hyperref}
\usepackage[all]{hypcap}
\usepackage[nameinlink]{cleveref}

\usepackage[acronym,smallcaps,nowarn]{glossaries}
\glsdisablehyper{}
\usepackage{listings}
\lstdefinestyle{mystyle}{
    commentstyle=\color{OliveGreen},
    numberstyle=\tiny\color{black!60},
    stringstyle=\color{BrickRed},
    basicstyle=\ttfamily\scriptsize,
    breakatwhitespace=false,
    breaklines=true,
    captionpos=b,
    keepspaces=true,
    numbers=none,
    numbersep=5pt,
    showspaces=false,
    showstringspaces=false,
    showtabs=false,
    tabsize=2
}
\usepackage{centernot}

\crefname{lemma}{lemma}{lemmas}
\Crefname{lemma}{Lemma}{Lemmas}
\crefname{thm}{theorem}{theorems}
\Crefname{thm}{Theorem}{Theorems}
\crefname{prop}{proposition}{propositions}
\Crefname{prop}{Proposition}{Propositions}

\newcommand{\cN}{\mathcal{N}}

\newcommand{\g}{\, | \,}

\newcommand{\E}[2]{\mathbb{E}_{#1}\left[#2\right]}

\usepackage{booktabs,arydshln}
\makeatletter
\def\adl@drawiv#1#2#3{%
        \hskip.5\tabcolsep
        \xleaders#3{#2.5\@tempdimb #1{1}#2.5\@tempdimb}%
                #2\z@ plus1fil minus1fil\relax
        \hskip.5\tabcolsep}
\newcommand{\cdashlinelr}[1]{%
  \noalign{\vskip\aboverulesep
           \global\let\@dashdrawstore\adl@draw
           \global\let\adl@draw\adl@drawiv}
  \cdashline{#1}
  \noalign{\global\let\adl@draw\@dashdrawstore
           \vskip\belowrulesep}}
\makeatother

\newacronym{ELBO}{elbo}{evidence lower bound}
\newacronym{GMM}{gmm}{Gaussian mixture model}
\newacronym{KL}{kl}{Kullback-Leibler}
\newacronym{LDA}{lda}{latent Dirichlet allocation}
\newacronym{SVI}{svi}{stochastic variational inference}

\newacronym{MLE}{mle}{maximum likelihood estimate}
\newacronym{MCMC}{mcmc}{Markov chain Monte Carlo}
\newacronym{HMC}{hmc}{Hamiltonian Monte Carlo}

\newacronym{LBFGS}{l-bfgs}{limited-memory Broyden-Fletcher-Goldfarb-Shanno}
\newacronym{ADVI}{advi}{automatic differentiation variational inference}
\newacronym{NUTS}{nuts}{No-U-Turn sampler}

\newacronym{GLM}{glm}{generalized linear model}

\newacronym{IF}{if}{influence function}

\newacronym{PMF}{pmf}{Poisson matrix factorization}

\newacronym[\glsshortpluralkey={rpm}]
{RPM}{rpm}{reweighted probabilistic model}

\newacronym{NDCG}{ndcg}{normalized discounted cumulative gain}

\newacronym{MAP}{map}{mean average precision}

\usepackage{bm} 
\usepackage{arydshln}
\usepackage{longtable}% for long tables

\usepackage{booktabs}
\theorembodyfont{\upshape}
\theoremheaderfont{\scshape}
\theorempostheader{:}
\theoremsep{\newline}

\jmlrvolume{106}
\jmlryear{2019}
\jmlrworkshop{Machine Learning for Healthcare}

\title[Medical Deconfounder]{The Medical Deconfounder: Assessing
Treatment Effects with \newline Electronic Health Records}

\author{\Name{Linying Zhang} \Email{linying.zhang@columbia.edu} \\
      \addr Department of Biomedical Informatics, Columbia University\\
      New York, New York, USA 
      \AND
      \Name{Yixin Wang} \Email{yixin.wang@columbia.edu} \\
      \addr Department of Statistics, Columbia University\\
      New York, New York, USA 
      \AND
      \Name{Anna Ostropolets} \Email{anna.ostropolets@columbia.edu} \\
      \addr Department of Biomedical Informatics, Columbia University\\
      New York, New York, USA 
      \AND
      \Name{Jami J. Mulgrave} \Email{jami.mulgrave@columbia.edu} \\
      \addr Department of Biomedical Informatics, Columbia University\\
      New York, New York, USA 
      \AND
      \Name{David M. Blei} \Email{david.blei@columbia.edu} \\
      \addr Department of Statistics, Department of Computer Science, Columbia University\\
      New York, New York, USA 
      \AND
      \Name{George Hripcsak} \Email{george.hripcsak@columbia.edu} \\
      \addr Department of Biomedical Informatics, Columbia University\\
      New York, New York, USA }

\everypar=\expandafter{\the\everypar\loosness=-1}

\begin{document}

\maketitle

\begin{abstract}

The treatment effects of medications play a key role in guiding
medical prescriptions. They are usually assessed with randomized
controlled trials (RCTs), which are expensive. Recently, large-scale
electronic health records (EHRs) have become available, opening up
new opportunities for more cost-effective assessments. However,
assessing a treatment effect from EHRs is challenging: it is biased by
\emph{unobserved confounders}, unmeasured variables that affect both patients' medical prescription and their outcome, e.g. the
patients' social economic status. To adjust for unobserved
confounders, we develop the \emph{medical deconfounder}, a machine
learning algorithm that unbiasedly estimates treatment effects from
EHRs. The medical deconfounder first constructs a substitute
confounder by modeling which medications were prescribed to each
patient; this substitute confounder is guaranteed to capture all
multi-medication confounders, observed or unobserved
\citep{wangandblei2018}. It then uses this substitute confounder to
adjust for the confounding bias in the analysis. We validate the
medical deconfounder on two simulated and two real medical data sets.
Compared to classical approaches, the medical deconfounder produces
closer-to-truth treatment effect estimates; it also identifies
effective medications that are more consistent with the findings in
the~medical~literature.
\end{abstract}

\clearpage

\section{Introduction}

The treatment effect of medications plays a key role in guiding medical
prescriptions. Usually, a treatment effect is assessed with a randomized controlled
trial (RCT): each patient is randomly assigned to the treatment or
the control group; only the treatment group receives the medication.
The treatment effect is then assessed by comparing the average outcome
of the two groups. RCTs are considered the gold standard for treatment
effect assessment~\citep{Concato2000RCT}. Their randomized treatment
assignments make RCTs immune to confounding bias and amenable to
classical statistical tests of
significance~\citep{Byar1976Randomization, Suresh2011Randomization}.
But, though theoretically sound, RCTs have substantial limitations: they
are expensive, labor-intensive, and time-consuming. Moreover, they also do not always
generalize to the real patient~population \citep{Deaton2018RCTgeneralizability, Sanson-Fisher2007RCTlimitations}.

Electronic health records (EHRs) have recently emerged as an appealing
alternative data source to RCTs for estimating treatment effects~\citep{Schuemie2018Empirical, Hripcsak2018laggedReg, Tannen2009EHR}. EHRs
contain large-scale observational data about the medical history of
patients, such as patient demographics, diagnosis, medications, and
laboratory tests. In particular, patients' medication records and
their lab tests can serve as evidence for medications' treatment
effect: we can view their medication records as the treatment
assignments and their lab tests as the outcome. This view of EHRs
opens up new opportunities for more cost-effective ways of estimating
treatment effects.

How can we use EHRs to estimate a treatment effect? A naive approach is to
compare, for each medication, the outcome of the treated and the
untreated patients. However, this approach leads to a biased
assessment of the treatment effect; the treated and untreated population
may not be comparable. For example, the two populations may be
different in their age distributions, and this difference in age can lead
to a difference in their health outcomes. Hence, naively comparing the
outcomes between the treated and the untreated does not lead to
correct treatment effect estimate of the medication. In causal
inference terms, age is a \emph{confounder}; it affects both whether a
patient is treated and her outcome. When confounders are
\emph{observed}, we can adjust for them using classical causal
inference methods like matching, subclassification and inverse
probability weighting~\citep{Imbens:2015, lopez2017estimation,
mccaffrey2013tutorial, zanutto2005using, rassen2011simultaneously,
lechner2001identification}.

However, in EHRs many confounders are \emph{unobserved}. For example,
a patient's social economic status (SES) can influence both what
medications she receives and her health condition. However, SES is an
integrated measure of a person's sociological (e.g., occupation and
education level) and economical (e.g., income) position in the
society; it is typically not recorded in EHR systems. Such unobserved
confounders challenge traditional causal inference methods;
these methods assume all confounders are observed \citep{hernan2019causal}.

To tackle this challenge, we develop the
\emph{medical deconfounder}, a machine learning approach that
unbiasedly assesses treatment effects from EHRs. The medical deconfounder
takes in patients' medication records (as the treatment) and lab tests
(as the outcome) from EHRs; it outputs a set of medications that are
deemed effective. To adjust for unobserved confounders, the medical
deconfounder first models patients' medication records using a
probabilistic factor model. It then constructs a substitute confounder
based on this probabilistic factor model; this substitute confounder
is guaranteed to capture all multi-medication confounders, both
observed and unobserved~\citep{wangandblei2018}. The medical
deconfounder finally fits an outcome model. This outcome model
describes how the lab test (outcome) depends on both the medications
prescribed and the substitute confounder. The dependence on
medications in the outcome model reflects the treatment effect of the
medications.

Why might the medical deconfounder work? The key idea is to infer
unobserved confounders by modeling how medications are prescribed
together. For example, consider a cohort of patients with type 2
diabetes mellitus. We are interested in which of the medications taken
by diabetic patients have an effect on their hemoglobin A1c (HbA1c).
One confounder is body mass index (BMI), which affects both the
medical prescription and the outcome HbA1c. If a patient is
overweight or obese (i.e. has a high BMI), they are often prescribed
with both diabetic medications and weight-lowering medications;
overweight or obese patients also have higher HbA1c. Moreover, BMI is
not recorded for all patient visits in the EHRs, rendering it an
\emph{unobserved confounder}. However, we can infer this unobserved
confounder---BMI---by looking at which medications are prescribed
together. If a patient is prescribed with both diabetic medications
and weight-lowering medications, she probably has a high BMI. This is
precisely what the medical deconfounder does; it constructs a
substitute for unobserved confounders by modeling which medications
are prescribed together.

In the next sections, we set up the treatment effect assessment
problem in causal inference notations. We then describe the medical
deconfounder and evaluate it on both simulation studies and real case
studies. We apply the medical deconfounder to four datasets: two
simulated and two real on distinct types of diseases. Across datasets,
the medical deconfounder produces closer-to-truth treatment effect
estimates than classical methods; it also identifies effective
medications that are more consistent with the medical literature.

\paragraph{Technical Significance} We propose the medical
deconfounder, a machine learning approach to treatment effect
estimation from EHRs. The medical deconfounder leverages probabilistic
factor models to improve treatment effect estimates from EHRs. Between
the two most popular options of probabilistic factor models (i.e.
Poisson matrix factorization (PMF) \citep{schmidt2009bayesian, gopalan2015scalable} and deep exponential family (DEF) \citep{ranganath2015deep}),
we find DEF helps to recover closer-to-truth treatment effects than
PMF.

% , and thus we propose to use DEF in the medical deconfounder.

\paragraph{Clinical Relevance} Assessing treatment effects is an
important task that guides medical prescription. However, this task
is challenging when the data comes from observational EHRs as opposed
to randomized experiments. The presence of multiple medications
further complicates the task. In this work, we propose the medical
deconfounder as a solution to treatment effect assessment with EHRs.
The medical deconfounder can adjust for unobserved confounders in EHRs
and identify medications that causally affects the clinical outcome of
interest.

\paragraph{Related work} This work draws on two threads of related
work.

The first body of related work is on probabilistic modeling for causal
inference. Probabilistic models excel at capturing hidden patterns of
high-dimensional data; examples include latent Dirichlet allocation
(LDA)~\citep{blei2003latent} and Poisson matrix factorization
(PMF)~\citep{schmidt2009bayesian, gopalan2015scalable}. Recently,
probabilistic modeling has been applied to causal inference. For
example, \cite{louizos2017causal} use variational autoencoders to
infer unobserved confounders from proxy variables.
\citet{kocaoglu2017causalgan} and \citet{Ozery-Flato2018adversarial}
connect generative adversarial network (GAN) and causal inference.
\citet{tran2017implicit}, \citet{wangandblei2018}, and
\citet{ranganath2018multiple} leverage probabilistic models for
estimating unobserved confounders of multiple causes. The medical
deconfounder in this work extends their use of probabilistic models
into assessing the treatment effect of medications.

The second body of related work is on multiple causal inference with
unobserved confounding. \citet{tran2017implicit} and
\citet{heckerman2018accounting} focus on genome-wide association
studies (GWAS); they consider single-nucleotide polymorphisms (SNPs)
as the multiple causes and estimate their effects on a trait of
interest (e.g., height). \citet{wangandblei2018} develop the
deconfounder algorithm for multiple causal inference; it leverages
probabilistic factor models to infer unobserved multi-cause
confounders from the assignments of the multiple causes. Multiple
causal inference with unobserved confounding was also studied in
\citet{ranganath2018multiple} with an information-theoretic approach;
their method is applied to estimate the causal effect of multiple lab
measurements on the length of stay in the ICU. More recently,
\citet{Bica2019timeseriesdeconf} extend the deconfounder algorithm to
time series data; they use recurrent neural network (RNN) to infer
time-dependent unobserved confounders for multiple causal inference.
The medical deconfounder presents another extension of the
deconfounder algorithm; it extends the deconfounder to assess causal
effect of multiple medications in EHRs.

% \red{yw / maybe cite existing works that assess causal effects in
% EHRs. e.g. those that use matching, ipw, subclassification and
% assume strong ignorability. The medical deconfounder extends them by
% allowing for unobserved confounders.}

\section{The medical deconfounder}

We frame treatment effect assessment as a multiple causal inference
\citep{wangandblei2018,ranganath2018multiple} and describe
the~medical~deconfounder.

\subsection{Treatment effect assessment as a multiple causal
inference} 

We first set up notation. Consider a dataset of $N$ patients and $D~(D
> 1)$ medications. Denote $\mathbf{A}_i$ as the medication record of
patient $i$, $i=1,
\ldots, n$; it is a binary vector of length $D$ that describes whether
patient $i$ has taken each of the $D$ medications~$\mathbf{A}_i =
(A_{i1}, \ldots, A_{iD}) \in \{0,1\}^D$. For example, the medication
record of patient $i$ is $\mathbf{A}_i=(0, 1, 0,
\ldots, 0)$ if she has only taken the second medication. Each patient
also has an outcome $Y_i$. For example, it can be the difference of
pre-treatment and post-treatment lab measurements of patient $i$. For
each patient, we observe both her medication records and her
outcome
\begin{align*}
\{(\mathbf{A}_i, Y_i):i=1, \ldots, n\}.
\end{align*}

% We use lower case $a_i$ and $y_i$ for the observed medications and
% outcome of patient $i$.

The goal of treatment effect assessment is to identify the medications
that (causally) affect the clinical outcome. In other words, all else
being equal, the clinical outcome of a patient should be different if
she had (or had not) taken the effective medication. We formulate this
goal as a (multiple) causal inference problem \citep{Imbens:2015,
rubin1974estimating, rubin2005causal,
wangandblei2018,ranganath2018multiple}. Denote $Y_i(\mathbf{a})$ as
the potential outcome of patient $i$ if she were assigned with
treatment $\mathbf{a}$ of the medications. Either factual or counterfactual, this treatment $\mathbf{a}$
is a $D$-dimensional binary vector of medications: $\mathbf{a} \in
\{0,1\}^D$. Then the $j$th medication causally affects the outcome if
the expected potential outcome of a patient is different had she taken
(or not taken) the $j$th medication:
\begin{align}
\label{eq:evalgoal}
\mathbb{E}\left[Y_i(A_{i1}, \ldots, A_{ij-1}, 1, A_{ij+1}, \ldots, A_D) -
Y_i(A_{i1}, \ldots, A_{ij-1}, 0, A_{ij+1}, \ldots, A_D)\right] \ne 0.
\end{align}
% {\color{red} when talking about regression coefficients in the
% empirical studies, we should say the coefficient is precisely
% calculating this difference.}

While treatment effects depend on all the potential outcomes
$\{Y_i(\mathbf{a}): \mathbf{a}\in \{0, 1\}^D\}$, we only observe one
of them---the one that corresponds to the patient's medication record:
$Y_i = Y_i(\mathbf{A}_i).$ To infer treatment effects from only the
observed data, we develop the \emph{medical deconfounder} by extending
the deconfounder algorithm for multiple causal
inference~\citep{wangandblei2018}. The deconfounder algorithm can
unbiasedly estimate $\mathbb{E}[Y_i(\mathbf{a})] -
\mathbb{E}[Y_i(\mathbf{a}')]$ for all $\mathbf{a}$ and $\mathbf{a}'$
(and hence the left hand side of \Cref{eq:evalgoal}). It assumes ``no
unobserved single-cause confounders'', i.e. no unmeasured variables
can affect the outcome and \emph{only one} medication
\citep{wangandblei2018}.

The idea of the deconfounder is to construct a substitute confounder
$Z_i$ by fitting a probabilistic factor model to the medication
records $\{\mathbf{A}_i: i=1, \ldots, n\}$. This constructed
substitute confounder $Z_i$ satisfies ignorability
\citep{rosenbaum1983central, imai2004causal}
\begin{align}
\label{eq:subignore}
Y_i(\mathbf{a}) \independent \mathbf{A}_i  \g   Z_i,
\end{align}
assuming ``no unobserved single-cause confounders.'' This ignorability
given $Z_i$ (\Cref{eq:subignore}) greenlights causal inference. We can
treat the substitute confounder $Z_i$ as if it were an observed
confounder and proceed with causal inference \citep{Imbens:2015}
\begin{align}
\label{eq:outcome}
\E{}{Y_i(\mathbf{a})} =
\E{Z}{\E{Y}{Y_i  \g  Z_i, \mathbf{A}_i = \mathbf{a}}}.
\end{align}
\Cref{eq:outcome} lets us conclude treatment effects from EHRs and
evaluate whether each medication is causally effective via
\Cref{eq:evalgoal}.

The medical deconfounder extends the deconfounder into medical
settings. It operates in two steps. First, we fit a probabilistic
factor model to all the medication records $\mathbf{A}_i$. This step
lets us construct a substitute confounder $Z_i$ for each patient. We
then fit an outcome model treating this substitute confounder $Z_i$ as
an observed confounder. The fitted outcome model leads to treatment
effect estimates of medications. We discuss the details of these two
steps in the next sections.

\subsection{The medical deconfounder}

We describe the medical deconfounder in details. We first discuss how
to construct a substitute confounder from prescription records in EHRs. Then we discuss how to
assess the treatment effect of medications with an outcome model.

\subsubsection{Constructing the substitute confounder}

\label{subsec:subconf}

\begin{figure}[]
  \centering 
  \includegraphics[width=3in]{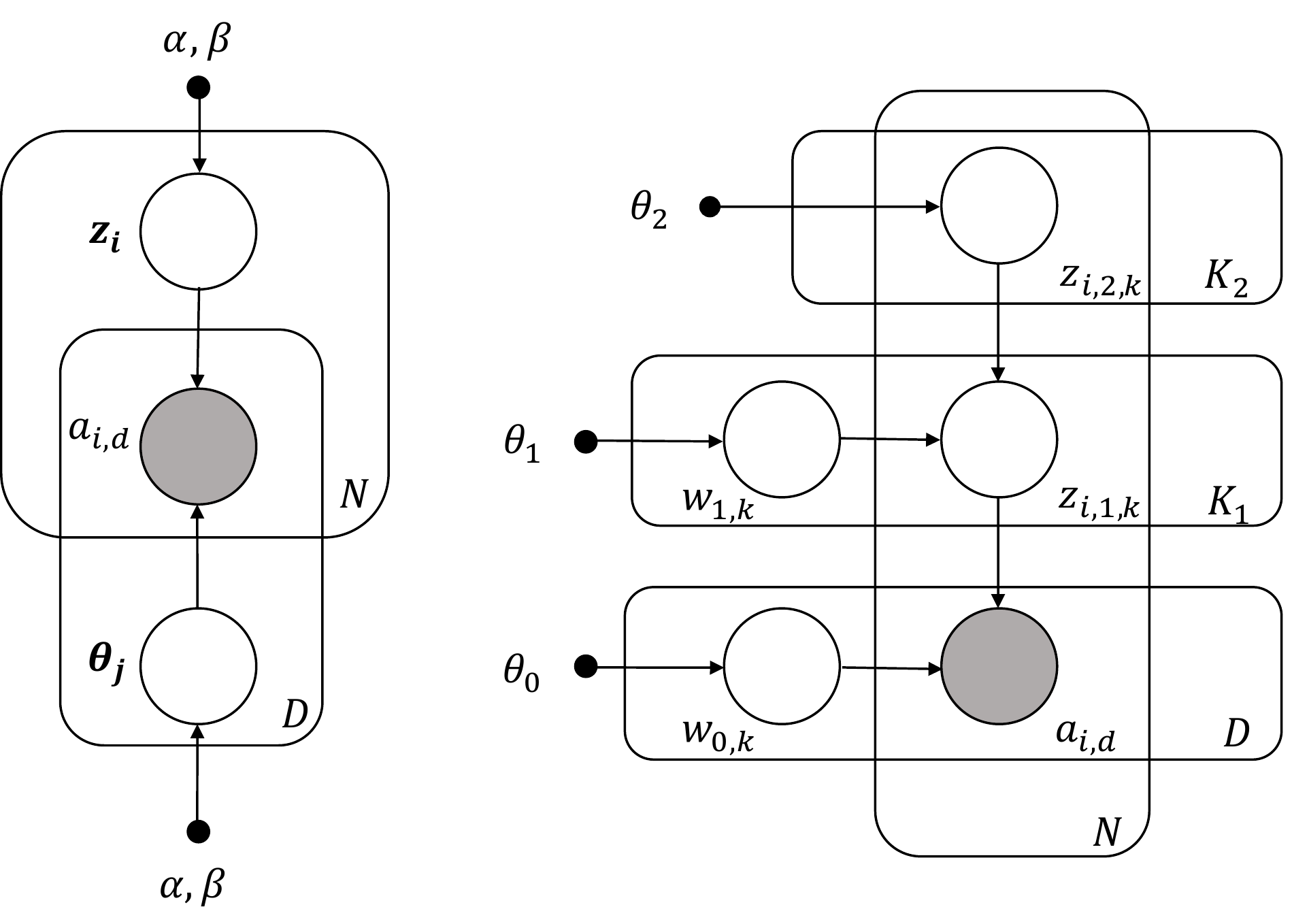}
  \caption{Graphical representation of PPCA (left), PMF (left) and a
  two-layer DEF (right). Random variables are represented by circles
  (shaded: observed; unobserved: hollow). Priors are represented by
  solid dots. In PPCA, $z_i$, $\theta_j$ and $a_{id}$ are modeled with
  normal distributions. In PMF, $z_i$ and $\theta_j$ are modeled with
  Gamma distribution, and $a_{id}$ are modeled with Poisson
  distribution.}
  \label{fig:graphs} 
\end{figure}

The medical deconfounder constructs a substitute confounder $Z_i$ by
fitting a probabilistic factor model of the medication records
$\{\mathbf{A}_i: i=1, \ldots, N\}$. This probabilistic factor model
needs to capture the observed distribution of the medication records
$p(\mathbf{A}_i)$. We study three options of the probabilistic factor
model for the medical deconfounder: probabilistic principal component
analysis (PPCA), Poisson matrix factorization (PMF), and deep
exponential family (DEF). \Cref{fig:graphs} shows the graphical
representations of the three probabilistic factor models.

\paragraph{Probabilistic principal component analysis (PPCA)} 
PPCA is a probabilistic formulation of PCA using a Gaussian latent
variable model \citep{tipping1999probabilistic}. For each patient $i$,
their medication record $\mathbf{A}_i = (A_{i1}, \ldots, A_{iD})$ is
modeled as a normal random variable; its mean is an inner product of a
$K$-dimensional latent variable $Z_i$ and some $(K\times
D)$-dimensional
parameter $\theta$; we posit a standard normal prior on each $Z_i$:
for $i=1, \ldots, N,$ and $j=1,\ldots, D,$ we have
\begin{align*}
Z_{i} &\sim \cN(0, \lambda^2),\\
A_{ij}  \g  Z_i &\sim \cN(z_i^T\theta_j, \sigma^2).
\end{align*}
The latent variable $Z_i$ will serve as the substitute confounder in
the medical deconfounder.

\paragraph{Poisson matrix factorization (PMF)}
PMF is a probabilistic factor model specific to modeling binary or
count data \citep{schmidt2009bayesian, gopalan2015scalable}. Because
each medication treatment $A_{ij}$ is binary---a patient either takes
or does not take a medication, we can model the patients' medication
records $\mathbf{A}_i$ with PMF. PMF is a similar factor model to PPCA
except in its distributional assumptions; PMF models each medication
treatment with a Poisson distribution and posits Gamma priors on the
latent variables $Z_i$: for $i=1, \ldots, N,$ and $j=1,\ldots, D,$
\begin{align*}
Z_i &\sim \textrm{Gamma}(\alpha,\beta),\\
\theta_j &\sim \textrm{Gamma}(\alpha, \beta),\\
A_{ij}  \g  Z_i&\sim\textrm{Poisson}(Z_i^T\theta_j).
\end{align*}
In PMF,  $Z_i$ is the patient-specific latent variable for patient $i$;
$\theta_j$ is medication-specific latent variable for medication $j$;
$A_{ij}$ indicates whether patient $i$ took medication $j$. Both $Z_i$
and $\theta_j$ are $K$-dimensional random variables. The latent
variable $Z_i$ will serve as the substitute confounder downstream.

\paragraph{Deep exponential family (DEF)} A DEF is a flexible
probabilistic factor model that has multiple layers of latent variables as in
neural networks \citep{ranganath2015deep}. We focus on a two-layer
DEF; it has the following structure:
\begin{align*}
W_{l,k} &\sim \textrm{Gamma}(\alpha, \beta),\\
Z_{i,2,k} &\sim \textrm{Gamma}(\alpha, \beta),\\
Z_{i,1,k}  \g  Z_{i,2,k} &\sim \textrm{Gamma}(\alpha, g(W_{1,k}Z_{i,2,k})),\\
A_{i,d}  \g  Z_{i,1} &\sim \textrm{Poisson}(g(W_0Z_{i,1})).
\end{align*}
The variable $Z_{i,l,k}$ corresponds to the $k$th latent
variable in the $l$-th layer for patient $i$. The variable $W_{l,k}$ is a
$K$-dimensional weight vector in the $l$-th layer. The variable $A_{id}$ is a
binary indicator of whether patient $i$ is prescribed with medication $j$.

In all three probabilistic factor models, the latent variable $Z_i$
will serve as the substitute confounder in downstream treatment effect
estimation. Specifically, we will fit the probabilistic model, i.e.
infer~$\theta$, using Markov chain Monte Carlo methods
\citep{robert2005montecarlo} or variational inference
\citep{jordan1999variational, blei2017variational}. We then compute the posterior expectation
of $Z_i$ given the inferred $\hat{\theta}$,
\begin{align*}
\hat{Z}_i\triangleq\E{Z}{Z_i \g  \mathbf{A}_i, \hat{\theta}}.
\end{align*}
If the probabilistic factor model fits the data well, then we can use
the constructed substitute confounder~$\hat{Z}_i$ in the downstream
treatment effect assessment.

To assess the adequacy of the probabilistic factor model, we follow
\citet{wangandblei2018} to perform a predictive check \citep{gelman1996posterior}. For each
patient $i$, we randomly hold out $s\%$ entries of her medication
record $\mathbf{A}_i$. The predictive check then proceeds in three
steps:
\begin{enumerate}
\item Generate replicated datasets for the heldout entries based on
the inferred posterior $p(Z_i  \g  \mathbf{A}_i, \hat{\theta})$.
\item Compare the value of a test statistic on the replicated
datasets to that of the observed dataset. The test statistic is the
expected log-likelihood of the heldout entries
\begin{align*}
t(\mathbf{X}_\textrm{heldout}) \triangleq
\mathbb{E}_{\mathbf{Z}, \theta}[{\log p(\mathbf{X}_\textrm{heldout}
\,  \g  \, \mathbf{Z}, \hat{\theta}) \,  \g  \,
\mathbf{X}_\textrm{obs}}].
\end{align*}
We compute this test statistic on both the observed dataset
$\mathbf{X}_{\textrm{heldout}}$ and each replicated dataset~$\mathbf{X}^\textrm{rep}_{\textrm{heldout}}$.
\item Conclude the probabilistic factor model is adequate if the
predictive score is close to 0.5. The predictive score is defined as
\begin{align}
\textrm{predictive score} \triangleq
p\left(t(\mathbf{X}^\textrm{rep}_{\textrm{heldout}}) <
t(\mathbf{X}_{\textrm{heldout}})\right).
\end{align}
A close-to-0.5 predictive score indicates neither under-fitting nor
over-fitting of the data \citep{wangandblei2018}. Otherwise, the
probabilistic factor model is inadequate.
\end{enumerate}
If a probabilistic factor model is deemed inadequate by the predictive
check, we must choose a different factor model. We repeat the
construction of the substitute confounder $\hat{Z}_i$ until one
constructed~$\hat{Z}_i$ passes the predictive check.

\subsubsection{Fitting a Bayesian linear regression outcome model} 

After constructing substitute confounder $\hat{Z}_i$, the medical
deconfounder adjusts for it as if it were an observed confounder in
causal inference. Specifically, we fit a Bayesian regression model to
$Y_i$ against both the medication record $\mathbf{A}_i$ and the
substitute confounder $\hat{Z}_i$
\begin{align*}
Y_i\sim \cN(\sum_{j=1}^D\beta_jA_{ij} + \sum_{k=1}^K \gamma_k \hat{Z}_{ik}, \sigma^2),
\end{align*}
where $K$ is the dimension of the substitute confounder $Z_i$. For studies with more than two medications, we posit an isotropic Gaussian prior $\cN(\textbf{0}, \alpha^{-1}\textbf{I})$ on all coefficients $\beta_j$ and $\gamma_k$.

We estimate the regression coefficients $\beta_j, j=1,
\ldots, D$ with mean-field variational inference. They indicate the
average treatment effect of each medication:
\begin{align}
\beta_j = \mathbb{E}\left[Y_i(A_{i1}, \ldots, A_{ij-1}, 1, A_{ij+1}, \ldots, A_{iD}) -
Y_i(A_{i1}, \ldots, A_{ij-1}, 0, A_{ij+1}, \ldots, A_{iD})\right].
\end{align}
When the coefficient $\beta_j$ is significantly different from zero,
we conclude that medication $j$ causally affects the clinical outcome
of interest.

\section{Simulation studies}

We first study the medical deconfounder on two simulation studies. The
two simulation studies are of distinct nature: one has only two
causes; the other has many causes. Below we first describe the
evaluation metrics and the baseline method we compare, and then
discuss the details of the two simulation studies.

\subsection{Performance metrics and baseline methods}

In both simulation studies, we evaluate the performance of the medical
deconfounder by the closeness-to-truth of its causal estimates. We
then compare these estimates with classical methods that do not adjust
for unobserved confounders.

\paragraph{Performance metrics} As a measure of closeness-to-truth in
simulations,  we compute the root mean square error (RMSE) between the
estimated treatment effects and the true effects. The RMSE is
defined as
\begin{align*}
    \mbox{RMSE}(\hat{\bm{\beta}}, \bm{\beta}) = \sqrt{\frac{1}{D}\sum_{j=1}^D (\hat{\beta}_j - \beta_j)^2},
\end{align*}
where $\hat{\bm{\beta}}$ is the estimated treatment effect and
$\bm{\beta}$ is the true effect.

We also evaluate the posterior distribution of the treatment effect by
``~\%~coverage,'' i.e. how often the estimated 95\% credible interval
(CI) covers the true treatment effect. We derive the 95\% credible
interval (CI) from the posterior distribution of the outcome model,
and compute the~\%~coverage by
\begin{align*}
    \mbox{\% coverage} &= \frac{\mathrm{Number\ of\ CI\ covers\ the\ truth}}{\mathrm{Number\ of\ total~treatments}} \times 100\%.
\end{align*}
% We compute the \% coverage for all medications, as well as separately for the causal medications and the non-causal medications.

\paragraph{Baseline methods} We compare the medical deconfounder with
classical methods that do not adjust for unobserved confounders. These
methods simply model the outcome as a function of the medical records
only; they do not adjust for any confounders. We call them "the
unadjusted model.'' Specifically, they fit the following Bayesian
regression model
\begin{align*}
    Y_i\sim \cN(\sum_{j=1}^D\beta_j A_{ij}, \sigma^2).
\end{align*}
They then take the $\beta$ coefficients as the effect size of each
medication.

In addition to the unadjusted model, we also compare the medical
deconfounder to an oracle model. The oracle model has access to the
true unobserved confounders $C_i$; it fits a Bayesian regression model
to both the medical records $\mathbf{A}_i$ (medications) and the true
confounders
\begin{align*}
    Y_i \sim \cN(\sum_{j=1}^D\beta_jA_{ij} + \sum_{k=1}^K
    \gamma_k C_{ik}, \sigma^2).
\end{align*}
We emphasize that these unobserved confounders $C_i$ are not available
in practice. The oracle model illustrates the best possible
performance in assessing treatment effects.

\paragraph{Computation} We fit probabilistic factor models using black
box variational inference \citep{ranganath2014black} as implemented in
Edward \citep{tran2016edward,tran2017deep}. We then draw 1000 samples
from the inferred posterior and fit the outcome model using automatic
differentiation variational
inference~(ADVI)~\citep{kucukelbir2017automatic} as implemented in the
\textrm{rstanarm} package \citep{carpenter2017stan} of R
\citep{team2013r}.

\subsection{Simulation study I: A two-medication simulation}

The first simulation study of the medical deconfounder is on a toy
example of only two medications. Under unobserved confounding, the
medical deconfounder is able to tell the causal (i.e.
causally-effective) medication from the non-causal (i.e.
non-causally-effective) medication. By contrast, the unadjusted model
returns both medications as causal.

\begin{figure}[h!]
  \centering 
  \includegraphics[width=3in]{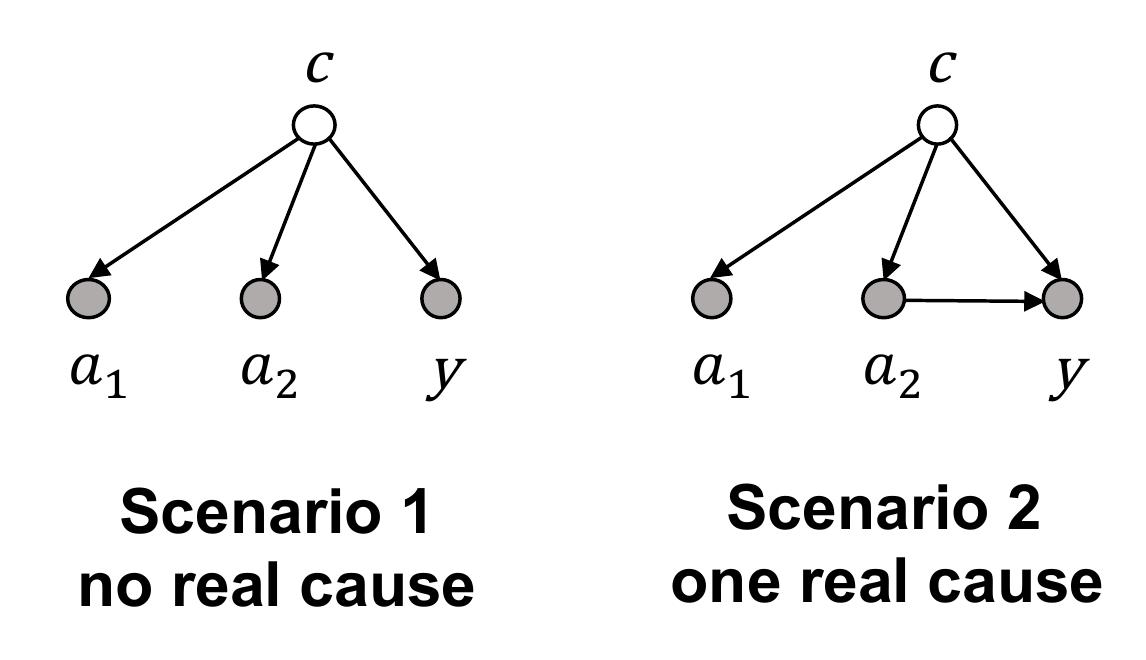}
  \caption{Causal graph of the two setups of the two-medication simulation study. Scenario 1 includes no real cause, while setup 2 has one real cause $a_2$. The confounder $c$ is a multi-medication confounder in both setups.}
  \label{fig:simulation} 
\end{figure}

\paragraph{Experimental setup}
We experiment the medical deconfounder in two setups. In both, there
is an unobserved confounder $C_i$ and two medications $A_{i1}$ and
$A_{i2}$ for each patient $i$. The unobserved confounder $C_i$ is
multi-medication; both medications $A_{i1}$ and $A_{i2}$ are linearly
dependent on the unobserved confounder $C_i$. We then simulate a
continuous outcome $Y_i$ that is also linearly dependent on the
confounder $C_i$. We consider two setups of the outcome. In the first
setup, neither of the causes is causal. In the second, one of the
causes is causal. (\Cref{fig:simulation} illustrates the two settings
with graphical models.)

Specifically, for each patient $i$, we simulate her confounder $C_i$
and the medication records $\mathbf{A}_i$ as
\begin{align*}
C_i &\sim \cN(0,1),\\
A_{1i} &= 0.3C_i + \epsilon_i, \\
A_{2i} &= 0.4C_i + \epsilon_i,
\end{align*}
where $\epsilon_i \sim \cN(0,1)$. In the first setup, the outcome is simulated as
\begin{align*}
Y_{i} = 0.5C_i + \epsilon_i.
\end{align*}
In the second, it is simulated as
\begin{align*}
Y_{i} = 0.5C_i
+ 0.3A_{2i} + \epsilon_i,
\end{align*}
where $\epsilon_i \sim \cN(0,1)$. In both setups, we simulate a sample
size of $N=1,000$.

% We simulate these two setups because we hope to see that the medical
% deconfounder is able to not only render medications non-causal after
% adjusting for the proxy of hidden confounder as in setup 1, but also
% preserve the true causes in setup 2 after adjustment.

\paragraph{Deconfounding with the medical deconfounder}  We use
probabilistic principal component analysis (PPCA) with latent
dimensionality $K=1$ as the probabilistic factor model. To assess the
model fit, we perform the predictive check for the factor model by
randomly holding out 20\% of the data. The fitted model returns a
predictive score close to 0.5; it passes the predictive check.

\paragraph{Results}
\begin{table}[h!]
\centering
\begin{tabular}{lllll}
				\toprule& \multicolumn{2}{c}{medication 1} & \multicolumn{2}{c}{medication 2} \\ \cline{2-5} 
             & \textbf{coef (std err)} &  \textbf{$p$-value} &  \textbf{coef (std err)} &  \textbf{$p$-value} \\ 
             \midrule
 \textbf{Truth}        & 0.000          & 1       & 0.000          & 1       \\ 
 % \hline
 \textbf{Oracle}       & 0.025 (0.039)  & 0.522   & -0.022 (0.040) & 0.594   \\ 
 \hdashline
 % \hline
 \textbf{Unadjusted}   & 0.125 (0.042)  & 0.003   & 0.141 (0.041)  & 0.001   \\ 
 % \hline
 \textbf{Med. Dcf.} & 0.020 (0.081)  & \textbf{0.803}   & 0.052 (0.071)  & \textbf{0.461}   \\     
				\bottomrule
\end{tabular}
 \caption{Estimated treatment effects in the two-medication simulation with no real cause. The $p$-value for each medication tests the null hypothesis that the coefficient is equal to zero (no causal effect). The
 medical deconfounder (``Med. Dcf.'') returns closer-to-truth
 $p$-values of the coefficients than the baseline method.}
 \label{table1} 
\end{table}

\begin{table}[h!]
\centering
\begin{tabular}{lllll}
				\toprule& \multicolumn{2}{c}{medication 1} & \multicolumn{2}{c}{medication 2} \\ \cline{2-5} 
             & \textbf{coef (std err)} &  \textbf{$p$-value} &  \textbf{coef (std err)} &  \textbf{$p$-value} \\ 
             \midrule
 \textbf{Truth}        & 0.000          & 1       & 0.300          & 0       \\ 
 % \hline
 \textbf{Oracle}       & 0.058 (0.038)  & 0.132   & 0.329 (0.039) & 0.000   \\
  % \hline
\hdashline
 \textbf{Unadjusted}   & 0.181 (0.040)  & 0.000   & 0.469 (0.040)  & 0.000   \\ 
 % \hline
 \textbf{Med. Dcf.} & 0.069 (0.063)  & \textbf{0.272}  & 0.333 (0.072)  & \textbf{0.000}   \\     
				\bottomrule
\end{tabular}
 \caption{Estimated treatment effects in the two-medication simulation
 with one real cause. The $p$-value for each medication tests the null hypothesis that the coefficient is equal to zero (no causal effect). The medical deconfounder (``Med. Dcf.'') returns
 closer-to-truth $p$-values of the coefficients than the baseline
 method.}
 \label{table2} 
\end{table}

Table \ref{table1} and Table \ref{table2} present the regression
coefficients and $p$-values of the three models in the two experimental setups. We
compare the unadjusted model (no control), the medical deconfounder
(control for the substitute confounder), and the oracle model (control
for the true unobserved confounder). In both setups, the unadjusted
model leads to biased causal coefficient estimates. The medical
deconfounder reduces the bias of estimates, and returns causal
coefficients that are nearly the same as those from the oracle.

Moreover, the medical deconfounder is able to identify the true causal
medication in the second setup. After adjusting for the substitute
confounder, the coefficient of the true causal medication stays
significant while the non-causal one becomes insignificant. Their
$p$-values are consistent with whether they are causal. In contrast,
the unadjusted model returns statistically significant coefficients
for both medications; it leads to a wrong conclusion that both
medications are causal. In rare runs, the medical deconfounder did not
adjust the raw coefficient estimates significantly, but even then, it
increased the variance of the estimate of the non-causal medication so
that it can still correctly classify medications as causal or non-causal.

\subsection{Simulation study II: A multi-medication simulation}

We next evaluate the medical deconfounder on a multi-medication
simulated dataset. As in the first simulation, the medical
deconfounder improves the effect size estimates for the medications;
the confidence interval of treatment effect estimates also covers the
truth more often than classical methods.

\paragraph{Experimental setup} We simulate a dataset of $D=50$
medications and $N=5,000$ patients. The medication record
$\mathbf{A}_i$ of each patient is influenced by a ten-dimensional
multi-medication unobserved confounder $C_i$. A real-valued outcome is
simulated as a function of the confounder $C_i$ and the medication
record $\mathbf{A}_i$. The simulated dataset is at a similar scale to
the dataset we use in the empirical studies.

We simulate each multi-medication confounder $C_{ik}$ from a standard
normal distribution,
\begin{align*}
    C_{ik} \sim \cN(0,1),\quad k = 1, \dots,10.
\end{align*}
Then we simulate the medication record of each patient $i$ from a
Bernoulli distribution,
\begin{align*}
    A_{ij} \sim \mbox{Bern}(\sigma(\sum_{k=1}^K \lambda_{kj} C_{ik})), \quad j = 1, \dots, 50,
\end{align*}
where $\sigma(\cdot)$ is the sigmoid function and $\lambda_{kj} \sim
\cN(0, 0.5^2)$. Finally, we simulate a continuous outcome $Y_i$ as a
function of both the confounder and the medication record,
\begin{align*}
    Y_i = \sum_{j=1}^{D}\beta_j A_{ij} + \sum_{k=1}^{K}\gamma_k C_{ik} + \epsilon_i,
\end{align*}
where $\epsilon_i \sim \cN(0,1)$, $\beta_j \sim \cN(0, 0.25^2)$, and
$\gamma_k \sim \cN(0, 0.25^2)$. To mimic the sparsity of causal medications
in practice, we randomly select 80\% of the medications and set their
coefficients $\beta_j$ to zero, therefore, only 10 medications are causal.

\paragraph{Deconfounding with the medical deconfounder} We implement
two probabilistic factor models PMF and DEF for the medical
deconfounder. The PMF passes the predictive check with $K=450$; the DEF
passes the predictive check with 30 and 4 latent variables in each
layer. Both factor models yield predictive scores close to 0.5.

\begin{table}[h!]
\centering
\begin{tabular}{lllll}
				\toprule 
			  & \textbf{RMSE} && \multicolumn{2}{l}{\textbf{\% Coverage}} \\ \cline{3-5} 
             &&  \textbf{All} &  \textbf{Causal} &  \textbf{Non-causal} \\ 
             \midrule
 \textbf{Oracle}        & 0.05          &78        & 50          & 85       \\ 
 % \hline
     \hdashline
 \textbf{Unadjusted}       & 0.14 & 38  & 30 & 40  \\ 
 % \hline
 \textbf{Med. Dcf. (PMF)}   & \textbf{0.12} & 38  & 30 & 40   \\ 
 % \hline
 \textbf{Med. Dcf. (DEF) } & 0.13  & \textbf{48}  & \textbf{40} & \textbf{50}   \\     
				\bottomrule
\end{tabular}
  \caption{ RMSE and \% coverage of CI of the multi-medication
  simulation. (Lower RMSE is better; higher \% coverage is better.)
  The medical deconfounder produces closer-to-truth causal estimates
  than the unadjusted model. The CI of estimates from DEF covers more
  true effects than the unadjusted. }
  \label{simulation_table}
\end{table}

\paragraph{Results} Table \ref{simulation_table} summarizes the causal
estimation results of the oracle model, the unadjusted model, and the
medical deconfounder with PMF and DEF as probabilistic factor models.
The medical deconfounder with both probabilistic factors produce less
biased effect estimates compared to the unadjusted model. Also, 48\%
of the CI's from DEF covers the truth, higher than the 38\% from the
unadjusted model. The increase of \% coverage by DEF is a consequence
of both correctly identifying more causal treatments, and decreasing
the false positives.

\section{Case studies}

We apply the medical deconfounder to two case studies on real datasets
of distinct disease cohorts. In both studies, the medical deconfounder
identifies causal medications that are consistent with the
medical literature. Below we discuss the two
disease cohorts and present the empirical results.

\subsection{Cohort extraction and evaluation methods}

In both case studies, we extract patient cohorts from the Columbia
University Medical Center database. The database contains
de-identified electronic health records standardized and stored
according to the Observational Health Data Science and Informatics
(OHDSI) format \citep{Hripcsak2016OHDSI}. We apply the medical deconfounder to each cohort.
Medical experts then perform literature reviews and evaluate the
results returned by the medical deconfounder.

\begin{figure}[h!]
  \centering 
  \includegraphics[width=0.8\textwidth]{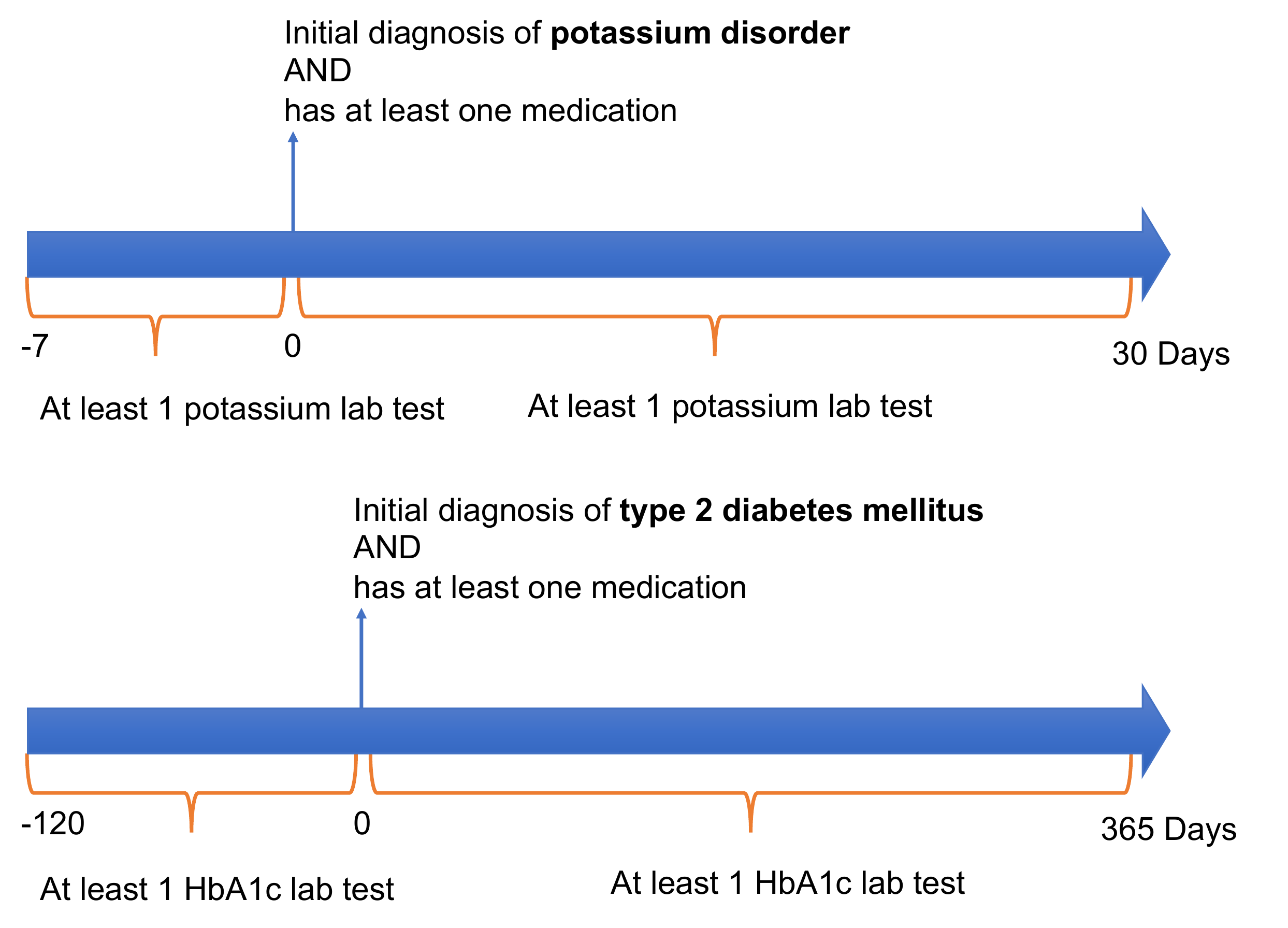}
  \caption{The diagram of cohort definition for potassium disorders
  (top) and type 2 diabetes mellitus (bottom). Patients meeting all
  criteria in the diagram are included in the cohort. Lines and arrows
  represent required intervals between events. }
  \label{fig:cohort} 
\end{figure} 

\paragraph{Case study I: Potassium disorders cohort.} Patients who meet
the following criteria are included in the potassium
disorder cohort:
\begin{itemize}
    \item was diagnosed with hypokalemia or hyperkalemia with
    continuous observation of at least 7 days before and 30 days after
    initial diagnosis (index date);
    \item has at least 1 measurement of potassium in serum/blood
    within 7 days prior to the first diagnosis;
    \item has at least 1 measurement of potassium in serum/blood
    within 30 days after the first diagnosis;
    \item has at least 1 medication exposure on the same day of
    initial diagnosis.
\end{itemize} 
After data preprocessing, there are 6185 patients and 33 unique
medications included in this cohort.

\paragraph{Case study II: Type 2 diabetes cohort.} Patients who meet
the following criteria are included in the type 2 diabetes
cohort:
\begin{itemize}
    \item  was diagnosed with type 2 diabetes with continuous
    observation of at least 30 days before and 30 days after the
    initial diagnosis (index date);
    \item has at least 1 measurement of HbA1C 120 days prior to the
    first diagnosis;
    \item has at least 1 measurement of HbA1C within 365 days after
    the first diagnosis;
    \item has at least 1 medication exposure on the same day of
    initial diagnosis.
\end{itemize}
After data preprocessing, there are 5564 patients and 30 unique
medications included in this cohort.

\paragraph{Data preprocessing} For both cohorts, patients' medication
records on the index date and their lab measurements immediately
before and after the index date are extracted from the database using
the OHDSI Atlas interface \citep{ohdsiAtlas}. All medications are
mapped to ingredients and dosage is ignored. To reduce the sparsity
of the patient-medication matrix, we remove the 5\% least frequent
ingredients from downstream analysis.

\paragraph{Evaluation methods}  Due to the unavailability of true
treatment effects in real datasets, we compare the medical
deconfounder estimates with the findings reported in the medical
literature. Medical experts perform literature review for all the
medications appeared in the studies; they look for evidence indicating
the presence or absence of causal relationships between the
medications and the outcome of interest.

\subsection{Case study I: Potassium disorders}

We apply the medical deconfounder to the patient cohort of potassium
disorders. Consider all the medications taken by the cohort of patients
with potassium disorders. The goal is to identify which of these
medications have causal effects on the serum potassium level. We find
that the medications identified to be causal by the medical
deconfounder are in concordance with the evidence from the medical
literature.

\begin{figure}
  \centering 
  \includegraphics[width=1.0\textwidth]{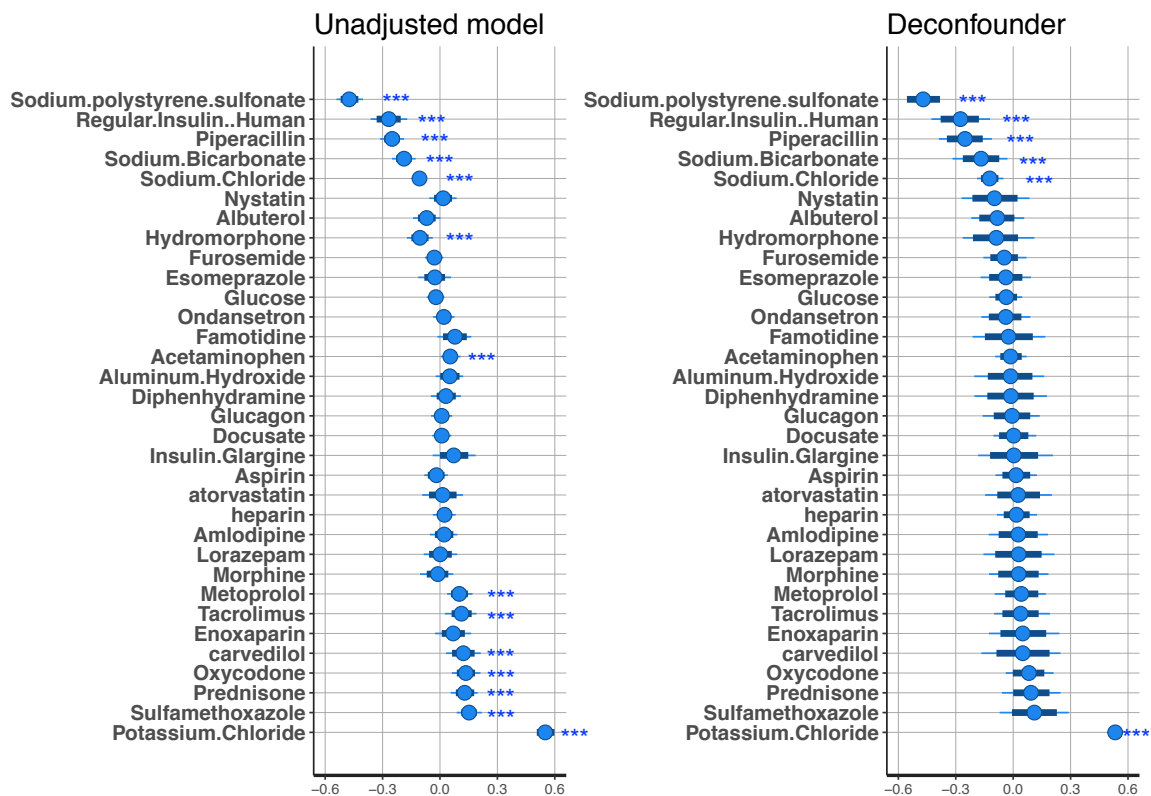}
  \caption{Treatment effects of medications in the potassium
  disorders cohort estimated by the unadjusted model (left) and the
  medical deconfounder (right). The medical deconfounder returns
  causal medications that are more consistent with the medical
  literature. The mean, 80\% credible interval, and 95\% credible
  interval of the estimated coefficients are indicated by the circle,
  the horizontal bar, and the solid line respectively. A medication is
  determined \emph{causal} if its 95\% credible interval excludes zero and is marked with "***". A positive
  coefficient means that the medication increases~serum~potassium.}
  \label{fig:potassium_result} 
\end{figure} 

\paragraph{Results} \Cref{fig:potassium_result} shows the coefficients
estimated by the medical deconfounder (control for the substitute
confounder) and the unadjusted model (no control). The medical
deconfounder reduces false positive discoveries while the true causal
medications remain significant after adjustment. (A medication is
determined \emph{causal} if its 95\% credible interval excludes zero.
)

Five medications are found to be causal by both models with well-supported medical literature on their physiological mechanisms: (1) sodium
polystyrene sulfonate is a potassium-binding resin commonly used to
treat hyperkalemia by increasing the excretion of potassium in stool
\citep{sps}; (2) insulin lowers serum potassium by internalizing
potassium intracellularly \citep{insulin2018hyperkalemia}; (3) piperacillin (often
prescribed with tazobactam) is a commonly used antibiotic for various
infections and are report to cause hypokalemia in a series of case
report \citep{pip1, pip2, pip3}; (4) sodium bicarbonate raises
systemic pH, a process accompanied by potassium movement into the
cells to maintain electroneutrality, leading to decrease of potassium
in the blood \citep{sodiumbicarb1, sodiumbicarb2}; (5) potassium
chloride is commonly administered to replenish potassium in patients
with low serum potassium.

Twenty-seven medications are identified as non-causal by the medical
deconfounder, including eight medications changing from causal to
non-causal after deconfounding. For most of these medications, we can
not find evidence in the medical literature that suggests their
influence on potassium, although a few medications may require more
detailed evaluation. Among them, one medication albuterol is reported
to have a potassium-lowering effect in patients with renal failure
\citep{albuterol}, but neither the unadjusted model nor the medical
deconfounder identifies it as a causal medication. We hypothesize that
this is because the cohort of renal failure patients in this dataset
is not large enough for this effect to be detected. The other medication,
furosemide, which is a diuretic used to reduce extra fluid in the body, has a delayed effect on potassium compared to other medications with immediate effect (e.g., sodium polystyrene sulfonate and regular insulin). Given this study uses the potassium measurement immediately after medications are prescribed to assess the treatment effect of all medications, there may not be enough time for the effect of furosemide to appear \citep{Mushiyakh2011Furosemide, furosemide}.

Two medications, changing from causal to
non-causal after deconfounding, are found to have an effect on potassium level in the literature. One medication is tacrolimus, which is an immunosuppressive medication
prescribed for patients with organ transplant to lower the risk of
organ rejection. Tacrolimus can increase serum potassium concentration due to
reduced efficiency of urinary potassium excretion
\citep{tacrolimus_K}. The other medication is sulfamethoxazole, which is an antibiotic to treat
infection. It is found to reduce renal potassium excretion through the
competitive inhibition of epithelial sodium channels when
co-administered with trimethoprim \citep{trimethoprim1,
trimethoprim2}.  These two medications are prescribed to patients with
relatively complicated health problems, and thus more scientific study
may be necessary to understand the mechanism. Even though the medical deconfounder does not identify these two medications to be causal but the unadjusted model does, the medical deconfounder still identifies effective
medications that are more consistent with the medical literature (six medications identified as causal only by the unadjusted model lack evidence for an effect on potassium).

%There are two medications, tacrolimus and sulfamethoxazole, found to
%have an effect on potassium level in the literature, are identified
%causal by the unadjusted model, but are determined to be non-causal
%after deconfounding.
\subsection{Case study II: Type 2 diabetes mellitus}

We next study the medical deconfounder on a patient cohort of type 2
diabetes mellitus. The goal is to identify medications that causally
affect hemoglobin A1c (HbA1c). HbA1c measures the percentage of a
protein called hemoglobin in the bloodstream that is bound by glucose;
it is a key indicator of the average blood glucose over the previous
two to three months \citep{Sherwani2016HbA1c}. In contrast to the
first case study where the treatment effect is immediate, HbA1c
reflects the long-term effect of medications on regulating blood
glucose. This long-term effect poses additional challenges in
treatment effect assessments.

\begin{figure}[h!]
  \centering 
  \includegraphics[width=0.8\textwidth]{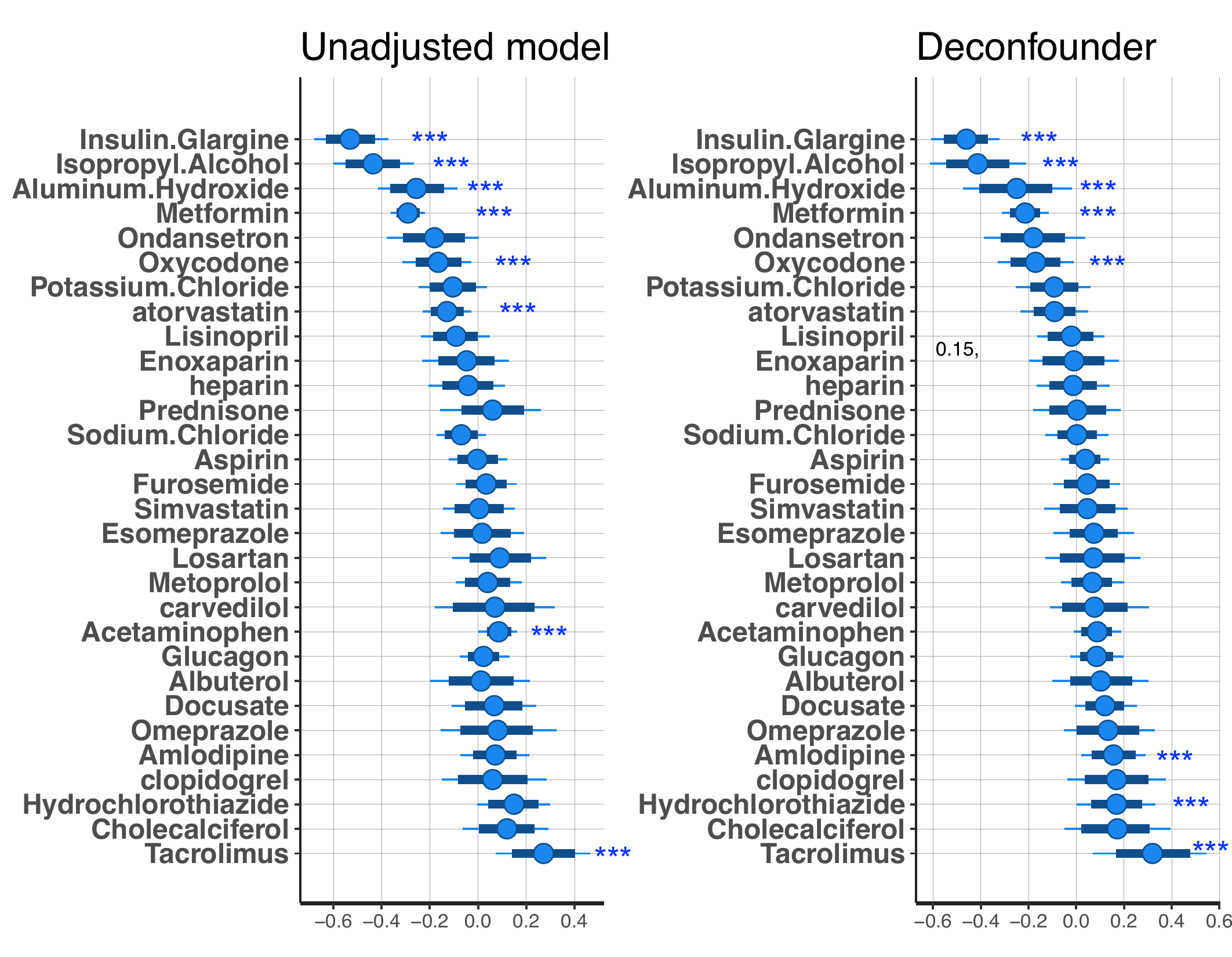}
  \caption{Treatment effects of medications in the diabetes cohort
  estimated by the unadjusted model (left) and the medical
  deconfounder (right). The medical deconfounder returns causal
  medications that are more consistent with the medical literature.
  The mean, 80\% credible interval, and 95\% credible interval of the
  estimated treatment effect are indicated by the circle, the
  horizontal bar and the solid line respectively. A medication is
  determined \emph{causal} if its 95\% credible interval excludes zero and is marked with "***". A negative treatment
  effect means that the medication down-regulates HbA1c, and a
  positive treatment effect means the medication up-regulates HbA1c. }
  \label{fig:t2dm} 
\end{figure} 

\Cref{fig:t2dm} shows the treatment effects estimated by the medical
deconfounder (control for the substitute confounder) and the
unadjusted model (no control).

The medical deconfounder returns three causal medications with
positive coefficients. Among the three, tacrolimus is the only
medication that is causal in both the medical deconfounder and the
unadjusted model. Both of the other two medications only appear
significant in the medical deconfounder. These two
medications---amlodipine and hydrochlorothiazide---are medications for
treating high blood pressure, a common comorbidity of diabetes. They
have been found to induce hyperglycemia in non-diabetic patients with
essential hypertension in several comparative studies
\citep{Fukao2011amlodipine, Cooper-DeHoff2010HCT}. These findings in
the literature are consistent with the positive treatment effect
estimates by the medical deconfounder. Moreover, both of the causal
medications are the first line recommended therapies for hypertension,
so the finding that the two medications can cause hyperglycemia are
important to guide the treatment decision of hypertension.

In more details, one of the medication amlodipine can induce
hyperglycemia likely because it blocks the calcium channels that
inhibits the release of insulin from $\beta$ cells in the pancreas
\citep{Sandozi2010Amlodipine}. The other medication
hydrochlorothiazide is a thiazide diuretics, a class of medications
that are known to promote hyperglycemia and in some cases contribute
to the new onset of diabetes \citep{Cooper-DeHoff2010HCT,
Gress2000HCT}. The exact mechanism is unknown, but it is postulated to
involve worsening of insulin resistance, inhibition of glucose uptake,
and decreased insulin release, among other pathways.

Two medications, acetaminophen and atorvastatin, are identified causal
by the unadjusted model, but are deemed non-causal in the medical
deconfounder. We do not find any evidence of causal relationship
between acetaminophen and blood glucose, except a few reports about
its interference on blood glucose sensors \citep{acetaminophen1, acetaminophen2}.
Atorvastatin is reported to increase the incidence of diabetes by
decreasing insulin sensitivity and increase ambient glycemia in
hypercholesterolemic patients \citep{atorvastatin}. Its estimated
effect by the unadjusted model is negatively causal. Although the
medical deconfounder is not able to identify this medication to be
causal with positive effect, the estimated treatment effect is more
positive after deconfounding, a change in the direction consistent
with its potential influence on increasing glucose.

The same five medications with a negative effect on HbA1c are returned
by both models. These include two well-known medications for treating
type 2 diabetes, insulin and metformin \citep{metformin1, metformin2,
insulin2009t2dm}. Isopropyl alcohol is not a medication but an ingredient in
alcohol-based sanitizers that are commonly used to clean patients'
skin before a blood test. A few studies were found addressing concerns
about the interference of isopropyl alcohol on the accuracy of blood
glucose test, but results are inconsistent among the studies
\citep{isopropylalcohol1, isopropylalcohol2}. There exists little
literature about aluminum hydroxide and oxycodone on their association
with blood glucose. These could be novel findings for further
investigations.

\section{Discussion} 

In this paper, we propose the medical deconfounder, a machine learning
algorithm for assessing treatment effects of medications with EHRs. For
a cohort of patients, the medical deconfounder works with multiple
relevant medications simultaneously and adjusts for unobserved
multi-medication confounders. The medical deconfounder then identifies
medications that causally affect the clinical outcome of interest. We
study the medical deconfounder on four datasets, two simulated and two
real. Across datasets, the medical deconfounder improves the treatment
effect estimates; it also identifies causal medications that are more
consistent with the medical literature than existing methods. These
empirical results show that the medical deconfounder can yield
insights around medication efficacy and adverse medication reactions.

As venues of future work, the medical deconfounder can be extended to
longitudinal settings, which will allow us to accommodate disease
progression and estimate time-dependent treatment effects of the
medications. We can also conduct sensitivity analyses of the treatment
effect estimates on the probabilistic factor model and the outcome
model. These analyses will allow us to understand how the modeling
choices in the medical deconfounder affect its treatment effect
estimates.

% ACKNOWLEDGEMENTS ONLY GO IN THE CAMERA-READY, NOT THE SUBMISSION
 \acks{This work was supported by NIH R01LM006910, NIH U01HG008680, ONR N00014-17-1-2131, ONR N00014-15-1-2209, NIH 1U01MH115727-01, NSF CCF-1740833, DARPA SD2 FA8750-18-C-0130, IBM, 2Sigma, Amazon, NVIDIA, and Simons Foundation.}

\bibliography{BIB1}

%\newpage
%\setcounter{table}{0}
%\renewcommand{\thetable}{A\arabic{table}}
%
% \section*{Appendix A. Tables of estimated treatment effects from
% empirical studies}
% \begin{table}[h!]
%  \centering 
%  \includegraphics[width=0.7\textwidth]{appendix_potassium_table_small}
%  \caption{Treatment effects of medications in the potassium
%  disorder cohort estimated by the unadjusted model and the medical
%  deconfounder. The mean, lower and upper bound of 95\% credible
%  interval of the estimated coefficients are included. Causal
%  medications found by each model are in bold; their 95\% credible
%  intervals exclude zero. A positive coefficient means
%  that the medication increases serum potassium and vice versa.}
%  \label{table:potassium_result_table} 
%\end{table} 
%
%\begin{table}[h!]
%  \centering 
%  \includegraphics[width=0.7\textwidth]{appendix_t2dm_table_small}
%  \caption{Treatment effects of medications in the diabetes
%  cohort estimated by the unadjusted model and the medical
%  deconfounder. The mean, lower and upper bound of 95\% credible
%  interval of the estimated coefficients are included. Causal
%  medications found by each model are in bold; their 95\% credible
%  intervals exclude zero. A negative coefficient means that the
%  medication decreases HbA1c and vice versa.}
%  \label{table:t2dm_result_table} 
%\end{table} 

\end{document}